\documentclass[10pt,twocolumn,letterpaper]{article}

\usepackage[pagenumbers]{cvpr} 

\usepackage{graphicx}
\usepackage{amsmath}
\usepackage{amssymb}
\usepackage{tabularx,booktabs}

\usepackage{bbding}
\usepackage{multirow}
\usepackage[table,xcdraw,dvipsnames]{xcolor}
\usepackage{enumitem}

\newcolumntype{Y}{>{\centering\arraybackslash}X}

\definecolor{cvprblue}{rgb}{0.21,0.49,0.74}
\usepackage[pagebackref,breaklinks,colorlinks,citecolor=cvprblue]{hyperref}

\def\paperID{2911} 
\def\confName{CVPR}
\def\confYear{2024}

\title{LION : Empowering Multimodal Large Language Model \\ with Dual-Level Visual Knowledge}
\author{
Gongwei Chen, Leyang Shen, Rui Shao\footnotemark[2], Xiang Deng, Liqiang Nie\footnotemark[2]\\
School of Computer Science and Technology, Harbin Institute of Technology, Shenzhen \\
 {\tt\small\{chengongwei, shaorui, dengxiang, nieliqiang\}@hit.edu.cn}\\
\texttt{\normalsize{\url{https://github.com/rshaojimmy/JiuTian}}}}

\begin{document}
\maketitle

\renewcommand{\thefootnote}{\fnsymbol{footnote}} 
\footnotetext[2]{Corresponding authors}
\begin{abstract}
Multimodal Large Language Models (MLLMs) have endowed LLMs with the ability to perceive and understand multi-modal signals. However, most of the existing MLLMs mainly adopt vision encoders pretrained on coarsely aligned image-text pairs, leading to insufficient extraction and reasoning of visual knowledge. To address this issue, we devise a dual-\textbf{L}evel v\textbf{I}sual kn\textbf{O}wledge e\textbf{N}hanced Multimodal Large Language Model (\textbf{LION}), which empowers the MLLM by injecting visual knowledge in two levels. \textbf{1)} \textbf{Progressive incorporation of fine-grained spatial-aware visual knowledge}. We design a vision aggregator cooperated with region-level vision-language (VL) tasks to incorporate fine-grained spatial-aware visual knowledge into the MLLM. To alleviate the conflict between image-level and region-level VL tasks during incorporation, we devise a dedicated stage-wise instruction-tuning strategy with mixture-of-adapters. This progressive incorporation scheme contributes to the mutual promotion between these two kinds of VL tasks. \textbf{2)} \textbf{Soft prompting of high-level semantic visual evidence}. We facilitate the MLLM with high-level semantic visual evidence by leveraging diverse image tags. To mitigate the potential influence caused by imperfect predicted tags, we propose a soft prompting method by embedding a learnable token into the tailored text instruction. Comprehensive experiments on several multi-modal benchmarks demonstrate the superiority of our model (\textit{e.g.}, improvement of 5$\%$ accuracy on VSR  and 3$\%$ CIDEr on TextCaps over InstructBLIP, 5$\%$ accuracy on RefCOCOg over Kosmos-2).
\end{abstract}

\section{Introduction}

\label{sec:intro}
\begin{figure}[t] 
	\centering
	\includegraphics[width=1\linewidth]{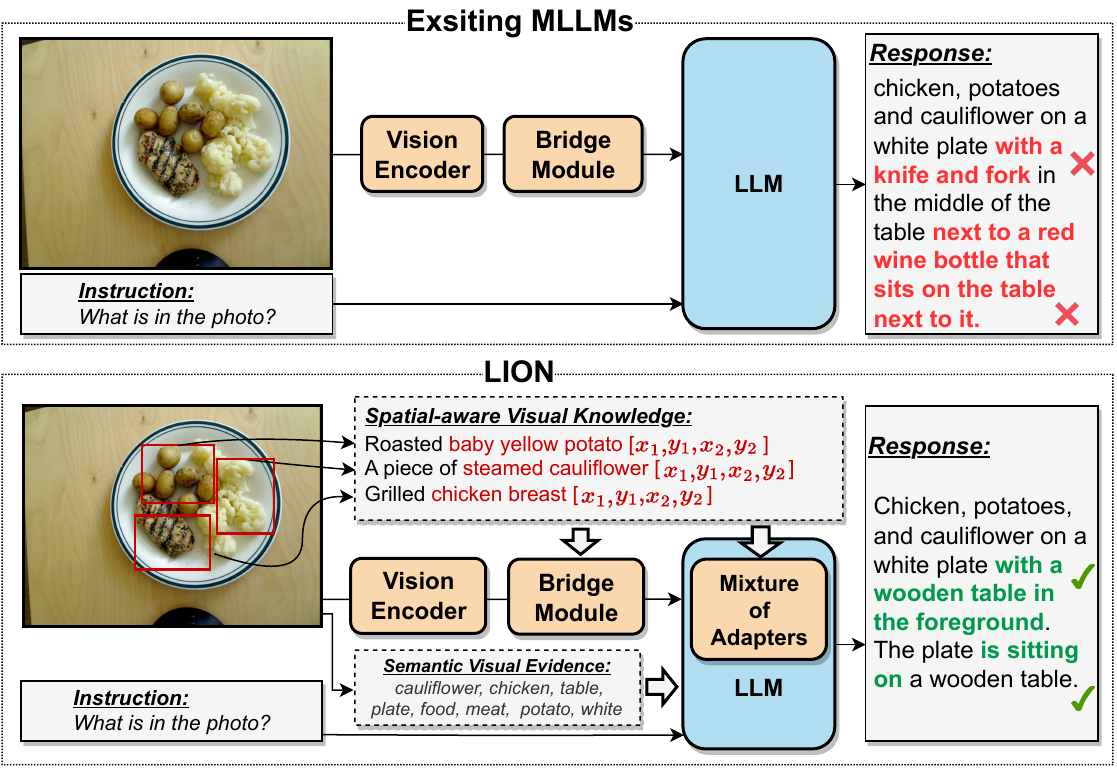}
	\caption{Comparison between existing MLLMs and LION \includegraphics[width=8pt]{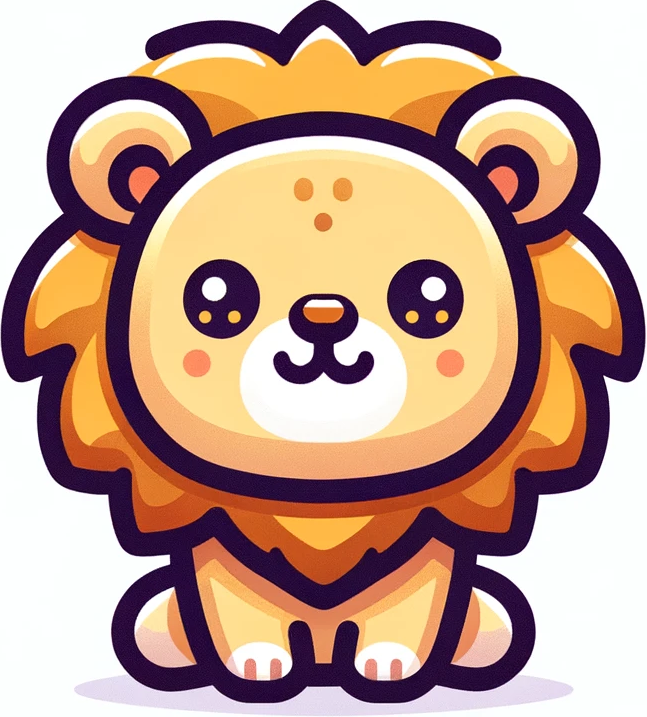}. The existing MLLM generates a vague and inaccurate response, while LION provides a more precise and contextually accurate description by progressively incorporating spatial-aware knowledge and softly prompting semantic visual evidence.}
	\label{fig:intro}
\end{figure}
Recently, Large Language Models (LLMs) have demonstrated remarkable zero-shot abilities on various linguistic tasks. Assisted by LLMs, several multimodal large language models (MLLMs), such as MiniGPT-4 \cite{zhu2023minigpt4}, Otter \cite{li2023otter}, and InstructBLIP \cite{dai2023instructblip}, achieve significant improvements in reasoning abilities to deal with various vision-language (VL) tasks. 

In most of the existing MLLMs, the visual information is mainly extracted from a vision encoder pretrained with image-level supervision (\textit{e.g.}, CLIP \cite{radford2021CLIP}), and then are adapted to a LLM by using a tiny bridge module. This makes these MLLMs inherently possess limited image understanding capabilities \cite{li2022grounded}. 
As shown in Fig.~\ref{fig:intro}, the insufficient visual information misleads MLLMs to provide erroneous and hallucinated responses. An intuitive solution to this problem is to replace or tune the vision encoder \cite{wang2023makes}. However, it requires pretraining on massive data or suffers from the catastrophic forgetting issue \cite{zhai2023catastrophicforget}, which diminishes the practical efficacy of this strategy. These predicaments highlight that the insufficient extraction of visual knowledge has become a central obstacle impeding the development of MLLMs.

To overcome this dilemma, as depicted in Fig.~\ref{fig:intro}, we devise a dual-\textbf{L}evel v\textbf{I}sual kn\textbf{O}wledge e\textbf{N}hanced Multimodal Large Language Model (\textbf{LION}), which enriches the visual information in MLLMs in two levels.
\textbf{1)} \textbf{\textit{Progressive incorporation of fine-grained spatial-aware visual knowledge}}.
LION enhances the MLLM with more fine-grained perceptual abilities by studying the region-level VL tasks involving the spatial coordinates. However, we find that simply training on region-level and the original image-level VL tasks\footnote{Here, image-level VL tasks denote image captioning and visual question answering, region-level VL tasks mean visual grounding tasks.} simultaneously hurts the general performances of the MLLM due to the conflicts between these two kinds of tasks. To address this issue, we propose a novel stage-wise instruction-tuning strategy to perform image-level and region-level VL tasks separately with two different visual branches and task adapters. In addition, we devise mixture-of-adapters with a router to dynamically fuse visual information across various granularities in a unified MLLM. This progressive incorporation of fine-grained visual knowledge contributes to the mutual promotion between these two kinds of VL tasks, and spawns LION to excel in capturing fine-grained visual information and performing spatial reasoning, as shown in Fig.~\ref{fig:intro}. \textbf{2)} \textbf{\textit{Soft prompting of high-level semantic visual evidence}}. Alongside the improvement of MLLMs in fine-grained perceptual capabilities, there is also an opportunity to enhance their high-level semantic understanding. LION uses an off-the-shelf vision model to extract high-level semantic knowledge, \textit{i.e.}, image tags, as supplementary information for the MLLM.
However, as off-the-shelf vision models are typically not flawless, errors in tag predictions are inevitable.
Inspired by prompt tuning, we propose a soft prompting method to mitigate the potential negative influence resulting from the imperfect predicted tags.
As shown in Fig.~\ref{fig:intro}, injection of semantic visual evidence alleviates the hallucination issue substantially.

\begin{figure}[h!]
    \centering
    \includegraphics[width=1\columnwidth]{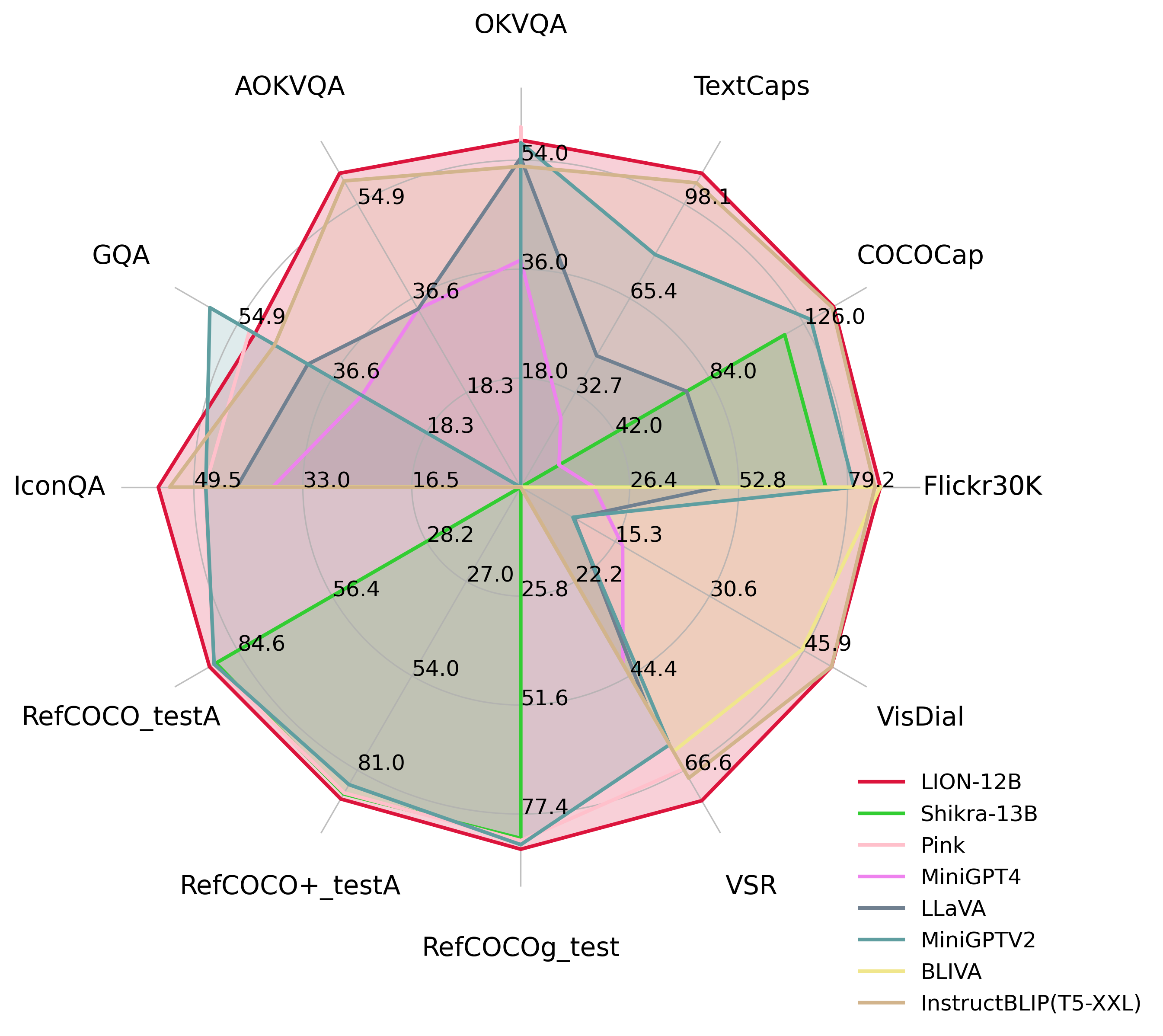}
    \caption{Compared to recently proposed MLLMs, LION achieves state-of-the-art performances across a wide range of VL tasks.}
    \label{fig:comp_polar}
\end{figure}

Our main contributions are summarized as follows:
\begin{itemize}[leftmargin=*]
     \item
     To address the internal conflict between region-level and image-level VL tasks, we propose a progressive incorporation of fine-grained spatial-aware visual knowledge with a novel stage-wise instruction-tuning strategy. It achieves mutual promotion between two kinds of VL tasks and equips LION with advanced holistic and fine-grained visual perceptual abilities.
    \item 
    As a powerful complement, we propose to integrate image tags as high-level semantic visual evidence into MLLMs, and design a soft prompting method to alleviate the bad influence from incorrect tags. This mitigates the hallucination issue and showcases positive effects on various VL tasks.
    \item 
    We evaluate LION on a wide range of VL tasks, including image captioning, visual question answering (VQA), and visual grounding, and demonstrate its superiority over the baselines as illustrated in Fig.~\ref{fig:comp_polar}.
    LION outperforms InstructBLIP by around 5$\%$ accuracy on VSR, and around 3$\%$ CIDEr on TextCaps, Kosmos-2 by around 5$\%$ accuracy on RefCOCOg. The evaluations on POPE and MMBench exhibit the remarkable abilities of LION in alleviating object hallucination and various perceptual dimensions.
\end{itemize}

\section{Related Work}
\label{sec:related work}

\subsection{Multimodal Large Language Models}
Building on the success of LLMs, many researches have emerged to extend them to multimodal tasks, especially VL tasks. 
The common pipeline uses a vision model to transform the image into visual features, followed by a bridge module to align them with the feature space of LLMs. Some works \cite{liu2023llava,zhang2023llamaAdapter,gao2023llamaAdapterV2,chen2023shikra,peng2023kosmos2} directly use a linear or MLP layer as bridge module, while others \cite{alayrac2022flamingo,zhu2023minigpt4,li2023blip2,dai2023instructblip,li2023otter,ye2023mplug} design more complicated bridge networks to compress or adaptively select visual information. Despite their impressive performance on VL tasks, there is still a lack of exploration on the effectiveness and limitation of the visual branch in a MLLM. Recently, Wang \textit{et al.} \cite{wang2023makes} empirically investigate factors contributing to the formation of an effective vision encoder in a MLLM from the perspective of pretraining. Differently, our work explores the effect of region-level VL tasks on the visual understanding abilities of the MLLM, and incorporates fine-grained and high-level visual knowledge to enrich the visual branch in the MLLM.

\begin{figure*}[t] 
	\centering
	\includegraphics[width=1\linewidth]{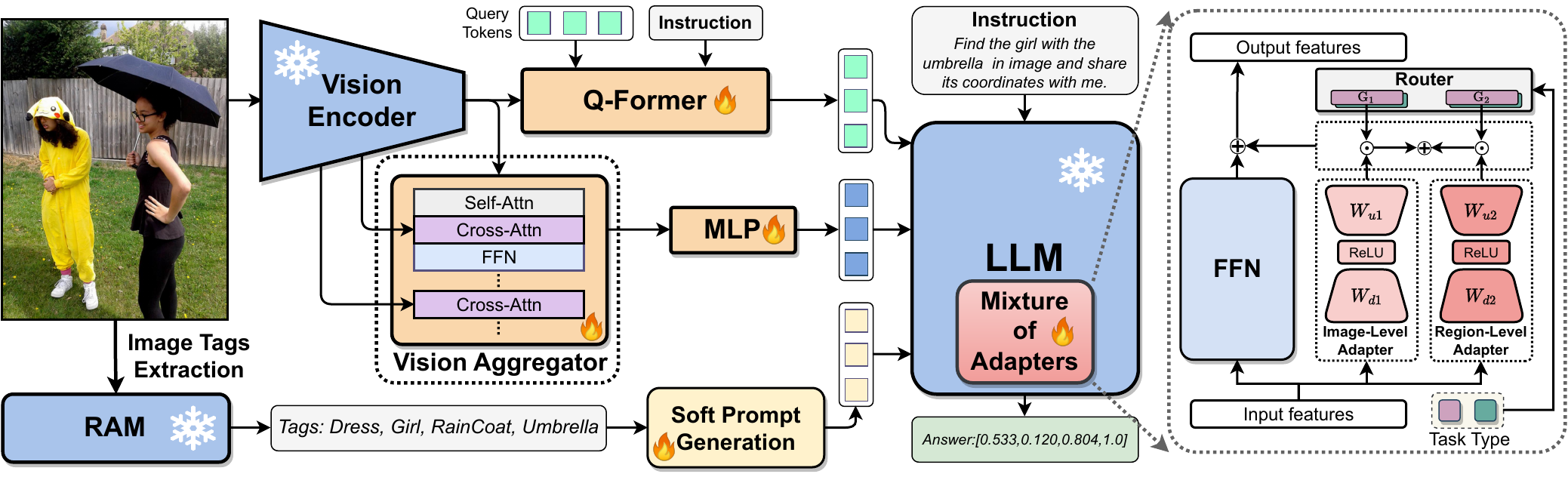}
    \caption{Overview of the proposed LION \includegraphics[width=8pt]{LION_logo.png}. The model extracts holistic visual features from Q-Former, and combines them with fine-grained spatial-aware visual features from the vision aggregator. The Mixture-of-Adapters with a router in the frozen LLM dynamically fuses visual knowledge learned from different visual branches and LLM adapters based on the task types (image-level and region-level).}
	\label{fig:framework}
\end{figure*}
\subsection{Visual Grounding in the field of MLLMs}

Visual grounding is a region-level VL task that aims to establish a connection between particular regions and their textual descriptors, which plays a vital role in human-machine interaction by enabling referential dialog. In the realm of MLLMs, there are some attempts to enhance MLLMs by leveraging visual grounding tasks. Works like Shikra~\cite{chen2023shikra}, Kosmos-2~\cite{peng2023kosmos2}, Ferret~\cite{you2023ferret} and Pink~\cite{xuan2023pink} demonstrate the promising direction of employing visual grounding datasets to endow MLLMs with region-level visual understanding abilities. They convert existing datasets equipped with spatial coordinates, like Visual Genome~\cite{krishna2017visual} and RefCOCO~\cite{kazemzadeh2014referitgame}, into the textual instruction format and perform instruction tuning on MLLMs. Merely considering the visual grounding task as one of several instruction-tuning tasks, these works fall short in exploring the interactions among various tasks. In contrast, our work investigates the internal conflict between visual grounding tasks and image-level VL tasks (\textit{e.g.}, image captioning and VQA), and proposes a stage-wise instruction-tuning strategy to address this issue, achieving a good balance between these two kinds of VL tasks.

\section{LION}
\label{sec:method}

In this section, we present the dual-Level vIsual knOwlEdge eNhanced multimodal large language model (LION). The proposed LION aims to enrich the visual information that is fed to the LLM in two ways, \textit{i.e.,} \textit{progressive incorporation of fine-grained spatial-aware visual knowledge} and \textit{soft prompting of high-level semantic visual evidence}. The whole framework is depicted in Fig.~\ref{fig:framework}.

\subsection{Progressive Incorporation of Fine-grained Spatial-Aware Visual Knowledge}
\subsubsection{Reorganizing Visual Grounding Tasks}
To incorporate fine-grained spatial-aware visual knowledge into MLLMs, we make use of region-level VL tasks, \textit{i.e.}, visual grounding, and meticulously process the data with spatial coordinates in a unified format for instruction-tuning MLLM. Visual grounding requires the model to generate or comprehend natural language expressions referring to particular objects or regions within an image, \textit{e.g.}, ``a man with glasses". Referring to objects or regions in complex images needs an ability of precisely comprehending fine-grained visual information. Current MLLMs lack such referring comprehending, as they mainly target a coarse alignment of VL modalities when pretrained on massive image-text pairs \cite{wang2023makes,chen2023pvit}. In this regard, we introduce visual grounding tasks as a kind of region-level VL tasks for the instruction-tuning of MLLMs. This aims to endow the model with fine-grained visual understanding ability such that better performance on image-level VL tasks (\textit{e.g.}, image captioning and VQA) might be achieved.

We adopt two types of visual grounding tasks, including referring expression comprehension (REC) and referring expression generation (REG) \cite{yu2016contextinREC}. We use the Visual Genome dataset \cite{krishna2017visualgenome}, which associates a local area with one short description, to construct REC/REG tasks. The templates used to organize the Visual Genome dataset in a unified instruction-tuning format can be found in \textbf{Appendix}.

One core point in reorganizing visual grounding tasks is the way of processing positions. Normally, the position of an object phrase is presented in the format of bounding box $[x_{min},y_{min},x_{max},y_{max}]$. We use a natural language style to describe object positions along with the square brackets. A sample in the REC task is displayed as follows: ``How can I locate a glass of beer in the image? Please provide the coordinates. Answer: $[0.525,0.0,0.675,0.394]$".

\subsubsection{The Stage-Wise Instruction-tuning Strategy}

\begin{figure}[t] 
    \centering
    \includegraphics[width=1\columnwidth]{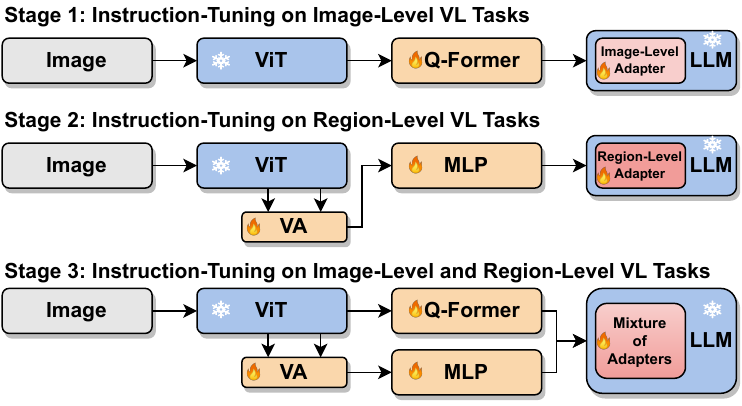}
    \caption{The stage-wise instruction-tuning strategy. \textbf{Stage 1}: We instruction-tune Q-Former and the image-level adapter on image-level VL tasks. \textbf{Stage 2}: We instruction-tune the vision aggregator (VA), MLP, and the region-level adapter on region-level VL tasks. \textbf{Stage 3}: The Mixture-of-Adapters is devised to form a unified model for instruction-tuning on both kinds of VL tasks.}
    \label{fig:stagewise_train}
\end{figure}
To facilitate MLLMs with fine-grained spatial-aware knowledge, the most intuitive way is to directly instruction-tune MLLMs with both image-level and region-level VL tasks in one stage. However, this single-stage instruction-tuning strategy is sub-optimal, and suffers from the internal conflict between these two kinds of VL tasks. We summarize two main issues leading to the internal conflict. \textbf{1) }One is the need of region-level modality-alignment pretraining. In concurrent works that integrate visual grounding ability, pretraining on the region-level multimodal datasets including visual grounding is a crucial step. Some works \cite{peng2023kosmos2,xuan2023pink,you2023ferret} elaborately create very large visual grounding datasets (\textit{e.g.}, GRIT-20M in kosmos-2 \cite{peng2023kosmos2}) to advance MLLM in fine-grained perception and understanding. The single-stage instruction-tuning makes it challenging to adapt visual representations learned for image-level alignment to region-level VL tasks under a limited training configuration. \textbf{2)} Another is the gap between the input-output modes of image-level VL tasks and region-level visual grounding tasks. The latter additionally requires MLLMs to understand specific phrases (in the format ``$[x_{min},y_{min},x_{max},y_{max}]$") about the positions of objects. They are semantically distinct from natural languages used in image-level tasks. This requirement necessitates the tuning of the LLM to adapt to region-level tasks, but may disrupt the internal state of the LLM suitable for image-level VL tasks. To address the above issues, we devise a stage-wise instruction-tuning strategy and mixture-of-adapters with a router.

The stage-wise instruction-tuning strategy is proposed to alleviate the internal conflict between image-level and region-level VL tasks during instruction-tuning. It is composed of three stages for instruction-tuning on image-level, region-level VL tasks and both, respectively, which is depicted in Fig.~\ref{fig:stagewise_train}. In stage 1, we follow instructBLIP \cite{dai2023instructblip} and fine-tune Q-Former and the image-level adapter \cite{chen2022adaptformer} in the LLM on image-level VL tasks, such as image captioning and VQA. In stage 2, we propose a vision aggregator for better capturing visual features in fine-grained understanding, which will be introduced later, and tune it with MLP and the region-level adapter on region-level VL tasks. The independent training in the first two stages greatly fulfills the requirements of sufficiently learning both image-level and region-level tasks, providing a solid foundation for subsequent joint training.

\paragraph{Mixture-of-Adapters with a Router.} In stage 3 of our stage-wise instruction-tuning, we need a unified model but encounter a situation where adapters of LLM in stages 1 and 2 are different and suit distinct input-output modes. Inspired by Mixture-of-Experts, we treat each adapter as an expert, and propose a router module to avoid the potential interference between them, as depicted in Fig.~\ref{fig:framework}.

An adapter \cite{chen2022adaptformer} is inserted at each FFN layer in a parallel manner. Assuming $X\in \mathcal{R}^{L\times D}$ is the hidden representations generated by a self-attention / causal attention layer, the output representations after FFN (represented as $\mathbf{F}$) with the adapter (denoted by $\mathbf{H}$) layer are formulated as,

\begin{equation}
    O = \mathbf{F}(X) + \mathbf{H}(X),
\end{equation}
\begin{equation}
    \mathbf{H}(X) = W_{u}(\sigma(W_{d}X)),
\end{equation}
where $\sigma$ is a non-linear function, ReLU. Our router module aims to dynamically aggregate the hidden features from the main branches and the multiple adapter branches according to task types. Given a set of adapters $\{\mathbf{H}_1, \dots, \mathbf{H}_K\}$, each kind of task $t$ defines a specific router function $\mathbf{R}^t$ to generate new hidden features, which can be formulated as,
\begin{equation}
    O^t = \mathbf{F}(X) + \sum^{K}_{k=1} \mathrm{G}^t_k\odot \mathbf{H}_k(X).
\end{equation}
where $\mathrm{G}^t_k \in \mathcal{R}^D$ is a trainable vector that modulates the hidden features from each adapter and makes them suitable for the target task. In practice, we define two types of tasks, one for image-level VL tasks (image captioning and VQA), the other for fine-grained VL tasks (visual grounding). Compared to directly incorporating multiple adapters, the router module provides a better way to maximize the complementarity of image-level and region-level tasks.

We use the standard language modeling loss in all instruction-tuning stages. In the experiments, we demonstrate that stage-wise training is superior to single-stage training, and ensures a good balance between high-level and fine-grained visual understanding capabilities, further achieves a significant mutual promotion between image-level and region-level VL tasks.
\subsubsection{Vision Aggregator}
To extract more sufficient visual details from input images, we devise a vision aggregator that integrates multi-level hidden features of the pretrained visual encoder. Although the vision encoder has a global reception field in all layers, it is verified that different transformer layers learn visual information at different scales \cite{dosovitskiy2020ViT}, \textit{e.g.}, lower layers learn visual details. Thus, our vision aggregator makes fine-grained spatial-aware visual knowledge more likely to be learned based on visual grounding tasks. Specifically, our vision aggregator can be regarded as a tiny transformer-style network, consisting of two transformer layers for aggregating the hidden features from the vision encoder. Given the hidden features $\{V_i,V_j,V_k\}$ from some middle layers in the vision encoder, the vision aggregation module uses two blocks to sequentially integrate the former two features with the last feature. Each block $\mathcal{B}$ is composed of self attention ($\mathrm{Attn}$), cross attention ($\mathrm{XAttn}$), and Feed-forward network ($\mathrm{FFN}$) arranged in a sequential manner. Finally, the output features $\bar{V}$ is generated as follows,
\begin{equation}
    \bar{V} = \mathcal{B}_2(\mathcal{B}_1(V_i;V_j);V_k),
\end{equation}
\begin{equation}
    \mathcal{B}(X;Y) = \mathrm{FFN}(\mathrm{XAttn}(\mathrm{Attn}(X),Y)).
\end{equation}
In practice, we use the middle layers $\{i=L-1, j=2L/3, k=L/3\}$ in the vision encoder to produce the hidden features as the input to VA, where $L$ is the number of layers in the vision encoder. 
\begin{figure}[t] 
    \centering
    \includegraphics[width=0.95\columnwidth]{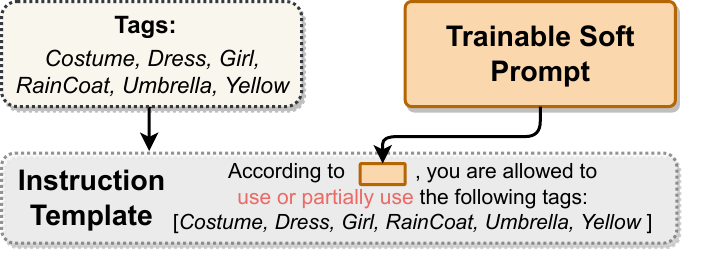}
    \caption{Instruction template with soft prompt. We use a well-designed instruction template with trainable soft prompts to inject the image tags generated by the RAM model into LION.}
    \label{fig:softprompt}
\end{figure}

\begin{table*}[t!]
\footnotesize
    \centering
    \caption{Comparison on image captioning and VQA. ``†" denotes including in-house data that are publicly inaccessible. ``*" means our evaluated results by using publicly released checkpoints, which are only for reference as official evaluation settings are incomplete. We report CIDEr score for Flickr30K, COCOCap, and TextCaps, Mean Reciprocal Rank (MRR) for Visual Dialog (VisDial), and top-1 accuracy for others. The best and second performances for each benchmark are indicated in bold and underline, respectively.}
    \begin{tabular}{l|ccc|cccccc}
    \toprule[1pt]
      Model & Flickr30K & COCOCap & TextCaps & OKVQA & AOKVQA & GQA & IconQA & VSR & VisDial \\
    \hline \hline
       Flamingo-3B \cite{alayrac2022flamingo} & 60.60  & 73.00 & - & - & - & - & - & - & 46.10 \\
       Flamingo-9B \cite{alayrac2022flamingo} & 61.50  & 79.40 & - & 44.70 & - & - & - & - & 48.00 \\
       Kosmos-1 \cite{huang2023kosmos1} & 67.10 & 84.70 & - & - &  & - & - & - & -  \\
       Kosmos-2 \cite{peng2023kosmos2} & 80.50 & - & - & - & - & - & - & - &  \\
       AdapterV2 \cite{gao2023llamaAdapterV2} & - & 122.20 & - & - & - & - & - & - & - \\
       Shikra \cite{chen2023shikra} & 73.90 & 117.50 & - & 47.16 & - & - & - & - & - \\
       Pink \cite{xuan2023pink} & - & - & - & \textbf{59.50} & - & \underline{52.60} & 47.80 & 66.30 & - \\
       MiniGPT4 \cite{zhu2023minigpt4} & \color{Gray}17.75* & \color{Gray}17.04* & \color{Gray}24.06* & 37.50 & \color{Gray}34.51* & 30.80 & 37.60 & 41.60 & \color{Gray}16.52* \\
       LLaVA \cite{liu2023llava} & \color{Gray}48.03* & \color{Gray}73.85* & \color{Gray}45.54* & 54.40 & \color{Gray}34.51* & 41.30 & 43.00 & 51.20 & \color{Gray}8.65* \\
       MiniGPTV2 \cite{chen2023minigptV2} & \color{Gray}80.75* & \color{Gray}129.16* & \color{Gray}80.60* & 56.90 & - & \textbf{60.30} & 47.70 & 60.60 & \color{Gray}8.47* \\
       BLIVA \cite{hu2023bliva} & \underline{87.10} & - & - & - & - & - & 44.88 & 62.20 & 45.63 \\
       \hline \hline
       \color{Gray}
       InstructBLIP† (T5XL) \cite{dai2023instructblip} & \color{Gray}84.50 & \color{Gray}138.21 & \color{Gray}82.55 & \color{Gray}49.28 & \color{Gray}57.86 & \color{Gray}48.40 & \color{Gray}50.00 & \color{Gray}64.80 & \color{Gray}46.60 \\
       \color{Gray}InstructBLIP† (T5XXL) \cite{dai2023instructblip} & \color{Gray}83.50 & \color{Gray}138.28 & \color{Gray}82.53 & \color{Gray}48.59 & \color{Gray}56.16 & \color{Gray}47.90 & \color{Gray}51.20 & \color{Gray}65.60 & \color{Gray}48.50 \\
       InstructBLIP (T5XL) \cite{dai2023instructblip} & 83.71 & 135.47 & 104.17 & 47.38 & 56.12 & 46.34 & 52.47 & 69.93 & 48.75 \\
       InstructBLIP (T5XXL) \cite{dai2023instructblip} & 85.79 & \underline{138.63} & \underline{105.44} & 53.02 & 59.38 & 47.74 & 53.18 & 68.46 & \underline{50.41} \\
       \hline
    \rowcolor[HTML]{faf0e6}
       \textbf{LION-4B} & 85.57 & 138.20 & 104.87 & 51.08 & \underline{59.98} & 49.50 & \textbf{54.91} & \underline{72.96} & 50.02 \\
    \rowcolor[HTML]{faf0e6}
       \textbf{LION-12B} & \textbf{87.12} & \textbf{139.25} & \textbf{108.76} & \underline{57.33} & \textbf{60.87} & 51.56 & \underline{54.89} & \textbf{73.77} & \textbf{50.42} \\
    \bottomrule[1pt]
    \end{tabular}
    \label{tab:zero-shot_cap_vqa}
\end{table*}

\begin{table*}[h!]
\footnotesize
    \centering
    \caption{Comparison on REC. ``Avg." means the average of top-1 accuracy over all the 8 evaluation sets.}
    \begin{tabular}{l|ccc|ccc|cc|c}
    \toprule[1pt]
       \multirow{2}{*}{Model}  & \multicolumn{3}{c|}{RefCOCO} & \multicolumn{3}{c|}{RefCOCO+} & \multicolumn{2}{c|}{RefCOCOg} & \multirow{2}{*}{Avg.}  \\
         & val & test-A & test-B & val & test-A & test-B & val & test & \\
    \hline
    \rowcolor[HTML]{EFEFEF}
    \multicolumn{10}{c}{\textit{Zero-shot Setting}} \\
    \hline
       Kosmos-2 \cite{peng2023kosmos2} & 52.32 & 57.42 & 47.26 & 45.48 & \textbf{50.73} & 42.24 & 60.57 & 61.65 & 52.21 \\
       GRILL \cite{jin2023grill} & - & - & - & - & - & - & - & 47.50 & - \\
       Pink \cite{xuan2023pink} & 54.10 & \textbf{61.20} & 44.20 & 43.90 & 50.70 & 35.00 & 59.10 & 60.10 & 51.00 \\
    \rowcolor[HTML]{faf0e6}
       \textbf{LION-4B} & 57.89 & 56.07 & 58.40 & \textbf{46.38} & 45.29 & 47.50 & 64.74 & 63.56 & 54.98 \\
    \rowcolor[HTML]{faf0e6}
       \textbf{LION-12B} & \textbf{58.54} & 56.41 & \textbf{59.36} & 45.93 & 45.73 & \textbf{47.89} & \textbf{66.12} & \textbf{64.69} & \textbf{55.58} \\
    \hline
    \rowcolor[HTML]{EFEFEF}
    \multicolumn{10}{c}{\textit{Fune-tuning Setting}} \\
    \hline
       OFA-L \cite{wang2022ofa} & 79.96 & 83.67 & 76.39 & 68.29 & 76.00 & 61.75 & 67.57 & 67.58 & 72.65 \\
       VisionLLM-H \cite{wang2023visionllm} & - & 86.70 & - & - & - & - & - & - & - \\
       Shikra-7B \cite{chen2023shikra} & 87.01 & 90.61 & 80.24 & 81.60 & 87.36 & 72.12 & 82.27 & 82.19 & 82.93 \\
       Shikra-13B \cite{chen2023shikra} & 87.83 & 91.11 & 81.81 & 82.89 & 87.79 & 74.41 & 82.64 & 83.16 & 83.96 \\
       Pink \cite{xuan2023pink} & 88.30 & 91.70 & 84.00 & 81.40 & 87.50 & 73.70 & 83.70 & 83.70 & 84.25 \\
       Ferret-7B \cite{you2023ferret} & 87.49 & 91.35 & 82.45 & 80.78 & 87.38 & 73.14 & 83.93 & 84.76 & 83.91 \\
       Ferret-13B \cite{you2023ferret} & 89.48 & 92.41 & 84.36 & 82.81 & 88.14 & 75.17 & \textbf{85.83} & \textbf{86.34} & 85.57 \\
       MiniGPTv2 \cite{chen2023minigptV2} & 88.69 & 91.65 & 85.33 & 79.97 & 85.12 & 74.45 & 84.44 & 84.66 & 84.29 \\ 
    \rowcolor[HTML]{faf0e6}
       \textbf{LION-4B} & 89.73 & 92.29 & 84.82 & 83.60 & 88.72 & 77.34 & 85.69 & 85.63 & 85.98 \\
    \rowcolor[HTML]{faf0e6}
       \textbf{LION-12B} & \textbf{89.80} & \textbf{93.02} & \textbf{85.57} & \textbf{83.95} & \textbf{89.22} & \textbf{78.06} & 85.52 & 85.74 & \textbf{86.36} \\
    \bottomrule[1pt]
    \end{tabular}
    \label{tab:ana_REC}
\end{table*}

\subsection{Soft Prompting of High-Level Semantic Visual Evidence}
The vision encoder in a MLLM may be insufficient in comprehensively extracting visual information required by complex multi-modal tasks, although it has been trained on large-scale image-text pairs. It has been demonstrated that increasing the amount and quality of pretraining multimodal datasets can significantly improve the visual understanding capability of MLLM \cite{wang2023makes}, but inevitably induces prohibitive computational overhead. An appealing alternative is to harness the convenient and powerful off-the-shelf vision models to capture various aspects of visual content within an image as a supplement.

We choose the recognize anything model (RAM) \cite{zhang2023ram} as an off-the-shelf vision model to provide diverse tags, encompassing objects, scenes, actions, and attributes, as visual evidence to support comprehensive visual perception. Instead of directly adding tags in the instruction, we design a soft prompting method to guide the model to adaptively use the inserted tags in order to avoid the potential negative influence caused by the imperfect predictions from RAM.

In Fig.~\ref{fig:softprompt}, we present the instruction template of tags along with the soft prompt that is a trainable vector. Our soft prompting approach can be regarded as a kind of prompt tuning methods, which guides the model toward the right direction. In standard prompt tuning works, the right direction is directly formulated as the optimization for task goals. In our work, the right direction is specified by a tailored sentence, \textit{``According to $<$hint$>$, you are allowed to use or partially use the following tags:"}, and ``$<$hint$>$" will be replaced by the soft prompt. Our soft prompting method for inserting tags has some distinct properties. It is designed to adaptively select valuable information from tags, rather than serving a specific task, as seen in standard prompt tuning methods. Our method directly uses the output labels from a small off-the-shelf vision model to incorporate high-level semantic visual evidence into a MLLM, so as to eliminate extra computational overhead of the feature alignment.

\begin{table}
\footnotesize
    \centering
    \caption{The comparison of various strategies in the instruction-tuning period. ``REC Avg." represents the average score of all REC tasks. ``Held-in" denotes the average score of COCOCap, TextCaps, OKVQA, and AOKVQA, while ``Held-out" means the average score of Flickr30K, GQA, IconQA, VSR, and VisDial.}
    \begin{tabular}{l|cc|c}
    \toprule[1pt]
       \multirow{2}{*}{Strategy}  & \multicolumn{2}{c|}{Image-Level} & Region-Level \\
        & Held-in & Held-out & REC Avg. \\ \hline
       Single Stage  & 84.79 & 60.20 & 3.78 \\
       Stage-wise  & \textbf{88.07} & 61.91 & 54.46 \\
       \ \ w/ Router & 87.67 & \textbf{62.16} & \textbf{54.98} \\
       \bottomrule[1pt]
    \end{tabular}
    \label{tab:comp_stage_training}
\end{table}

\begin{table*}[h!]
\footnotesize
    \centering
    \caption{Ablation studies of dual-level visual knowledge. ``VG" means visual grounding tasks. ``Held-in" and ``Held-out" denote the training images of tasks are seen and unseen, respectively. ``REC Avg." means the average score of all REC tasks.}
    \begin{tabular}{cc|cccc|ccccc|c}
    \toprule[1pt]
    \multicolumn{2}{c|}{Components} & \multicolumn{4}{c|}{Held-in} & \multicolumn{5}{c|}{Held-out} & \multirow{2}{*}{REC Avg.} \\
        VG & Tags & COCOCap & TextCaps & OKVQA & AOKVQA & Flickr30K & GQA & VSR & IconQA & VisDial & \\
    \hline
       $\times$ & $\times$  & 135.47 & 104.17 & 47.38 & 56.12 & 83.71 & 46.34 & 69.93 & 52.47 & 48.75 & - \\
       \checkmark & $\times$ & 137.87 & 104.84 & 51.07 & 56.90 & 83.99 & 49.22 & \textbf{73.20} & 54.67 & 49.70 & 54.98 \\
       \checkmark & \checkmark & \textbf{138.20} & \textbf{104.87} & \textbf{51.08} & \textbf{59.98} & \textbf{85.57} & \textbf{49.50} & 72.96 & \textbf{54.91} & \textbf{50.02} & 54.92 \\
    \bottomrule[1pt]
    \end{tabular}
    \label{tab:ablation_study}
\end{table*}

\begin{table*}[h!]
\footnotesize
    \centering
    \caption{Evaluation of object hallucination on POPE benchmark. F1 score is the major metric for halluciantion evaluation.}
    \begin{tabular}{c|l|>{\columncolor[HTML]{faf0e6}}c|ccccc}
    \toprule[1pt]
        Datasets & Metrics & LION & Shikra \cite{chen2023shikra} & InstructBLIP \cite{dai2023instructblip} & MiniGPT-4 \cite{zhu2023minigpt4} & LLaVA \cite{liu2023llava} & mPLUG-Owl \cite{ye2023mplug} \\
        \hline
        \multirow{5}{*}{Random} 
         & \textbf{F1-Score} $\uparrow$ & 88.33 & 86.19 & \textbf{89.27} & 80.17 & 66.64 & 68.39  \\
         & Accuracy $\uparrow$ & \textbf{88.97} & 86.90 & 88.57 & 79.67 & 50.37 & 53.97  \\
         & Precision $\uparrow$ & \textbf{97.12} & 94.40 & 84.09 & 78.24 & 50.19 & 52.07  \\
         & Recall $\uparrow$ & 81.00 & 79.27 & 95.13 & 82.20 & 99.13 & 99.60 \\
         \hline
        \multirow{5}{*}{Popular} 
         & \textbf{F1-Score} $\uparrow$ & \textbf{85.94} & 83.16 & 84.66 & 73.02 & 66.44 & 66.94  \\
         & Accuracy $\uparrow$ & \textbf{86.77} & 83.97 & 82.77 & 69.73 & 49.87 & 50.90  \\
         & Precision $\uparrow$ & \textbf{91.69} & 87.55 & 76.27 & 65.86 & 49.93 & 50.46  \\
         & Recall $\uparrow$ & 80.87 & 79.20 & 95.13 & 81.93 & 99.27 & 99.40  \\
         \hline
        \multirow{5}{*}{Adversarial} 
         & \textbf{F1-score} $\uparrow$ & \textbf{84.71} & 82.49 & 77.32 & 70.42 & 66.32 & 66.82  \\
         & Accuracy $\uparrow$ & \textbf{85.37} & 83.10 & 72.10 & 65.17 & 49.70 & 50.67  \\
         & Precision $\uparrow$ & \textbf{88.69} & 85.60 & 65.13 & 61.19 & 49.85 & 50.34  \\
         & Recall $\uparrow$ & 81.07 & 79.60 & 95.13 & 82.93 & 99.07 & 99.33  \\
    \bottomrule[1pt]
    \end{tabular}
    \label{tab:obj_hallucination}
\end{table*}

\section{Experiments}
\label{sec:experiments}
In this section, we conduct extensive experiments to demonstrate the effectiveness of our model along with the quantitative and qualitative analyses. Please refer to \textbf{Appendix} for implementation details and training details.

\subsection{Evaluations on Image-Level VL Tasks}
Here, we evaluate multi-modal understanding abilities of LION on two kinds of image-level VL tasks, image captioning and VQA. Image captioning requires the model to generate a text description of the input image. We use COCO caption \cite{chen2015cococap}, TextCaps \cite{sidorov2020textcaps} and Flickr30K \cite{young2014flickr30k} as benchmarks, and report CIDEr as the evaluation metric. We utilize greedy search for caption generation. VQA provides an image along with a specific question for the model, asking for the output as an answer. We evaluate LION on six VQA datasets, including OKVQA \cite{marino2019okvqa}, AOKVQA \cite{schwenk2022aokvqa}, GQA \cite{hudson2019gqa}, IconQA \cite{lu2021iconqa}, Visual Spatial Reasoning \cite{liu2023vsr}, and Visual Dialog (VisDial) \cite{das2017visdial}. For OKVQA, A-OKVQA, and GQA, we employ an open-ended generation with a greedy decoding strategy, For IconQA, Visual Spatial Reasoning, and Visual Dialog, we match the output with various candidates, and select the candidate with the highest value as the prediction. We report Mean Reciprocal Rank (MRR) for Visual Dialog, and top-1 accuracy for other VQA tasks. The detailed descriptions of these datasets and inference instructions are presented in \textbf{Appendix}.

As shown in Table \ref{tab:zero-shot_cap_vqa}, LION achieves the best performance across 7 out of 9 benchmarks, and the second on the other 2 benchmarks. LION shares the same training datasets with InstructBLIP, except Visual Genome dataset adopted in our work and the in-house dataset, WebCapFilt, used in InstructBLIP. The amount of Visual Genome dataset (3.6M) is far smaller than WebCapFilt (14M). Compared to the original InstructBLIP trained with WebCapFilt, LION exhibits superior performances on all zero-shot evaluation benchmarks, showcasing a better generalization ability. We also re-implemented InstructBLIP on the same instruction-tuning datasets. The comparison shows the consistent and significant improvements of LION over the re-implemented InstructBLIP, demonstrating the effectiveness of incorporating dual-level visual knowledge. Shikra, MiniGPTV2 and Pink are also integrated with visual grounding abilities. Their inferior results on most image-level VL tasks exhibit that our proposed stage-wise instruction-tuning strategy and soft prompting of high-level semantic knowledge are very helpful in enhancing holistic visual understanding abilities of MLLMs.

\subsection{Evaluations on Region-Level VL Tasks}
To assess the fine-grained perceptual and reasoning abilities of LION, we evaluate it on three REC datasets, RefCOCO \cite{kazemzadeh2014referitgame}, RefCOCO+ \cite{kazemzadeh2014referitgame}, RefCOCOg \cite{mao2016refcocog}. REC requires the model to locate the target object given a referring expression. We follow the standard setting, and use accuracy as an evaluation metric, which means it is correct when the IOU between prediction and ground-truth is no less than $0.5$.

In Table \ref{tab:ana_REC}, we show the comparison between LION and other MLLMs with respect to the grounding abilities, under the settings of zero-shot and fine-tuning evaluations, respectively. In the zero-shot evaluation setting, we directly employ LION to generate coordinates of referring expressions on three datasets. Our model shows significant improvements on most evaluation sets over Kosmos-2 and Pink, except test-A sets of RefCOCO and RefCOCO+. The languages used in RefCOCOg are more flowery than those used in RefCOCO and RefCOCO+. The significant improvements on RefCOCOg clearly demonstrate that LION can handle complex referring expressions and has superior zero-shot spatial-aware visual understanding abilities.

In the fine-tuning setting, we fine-tune LION with training samples from three REC datasets. As shown in Table \ref{tab:ana_REC}, our model achieves the best performance on average and on most of the evaluation sets, indicating the advanced fine-grained perception ability of our model. Ferret proposes a spatial-aware visual sampler to handle free-form referred regions, and meticulously constructs an extensive grounding dataset with lots of efforts on data generation and filtering. However, LION can achieve superior performances compared to Ferret by using a simple vision aggregator and existing datasets, implying the effectiveness of fine-grained visual knowledge.

\begin{figure}[t] 
    \centering
    \includegraphics[width=1\columnwidth]{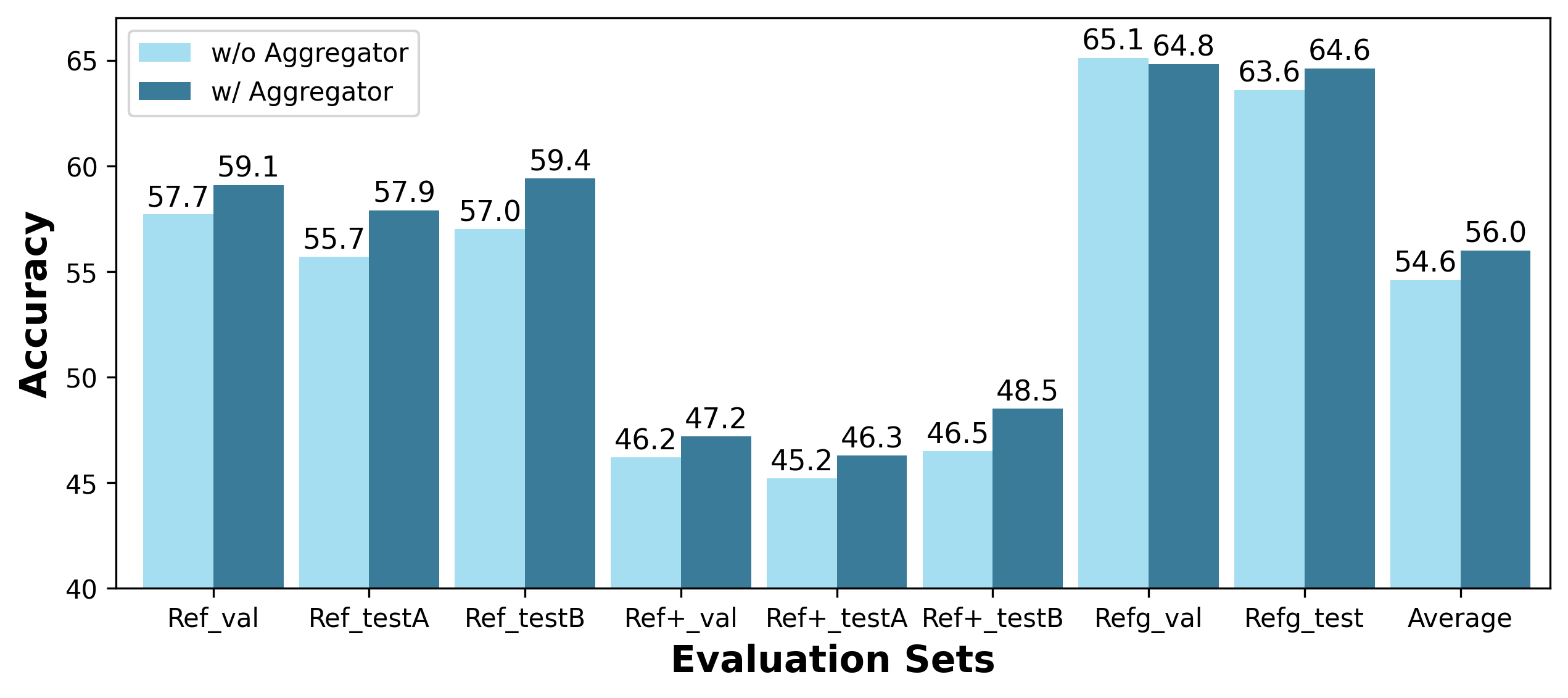}
    \caption{The effect of the vision aggregator. The results on RefCOCO, RefCOCO+, and RefCOCOg clearly show that the proposed module can overall improve REC performances across 8 evaluation sets. }
    \label{fig:effect_of_aggregator}
\end{figure}

\begin{table*}[h!]
\footnotesize
    \centering
    \caption{Evaluation on MMBench test set, all the reported results of compared models are from the leadboard of MMBench.}
    \begin{tabular}{c|c|c|c|cccccc}
       \toprule[1pt]
       Models & Text Encoder & Vision Encoder & Overall & LR & AR & RR & FP-S & FP-C & CP \\
       \hline
       MiniGPT-4~\cite{zhu2023minigpt4} & Vincuna-7B & EVA-G & 23.0 & 13.6 & 32.9 & 8.9 & 28.8 & 11.2 & 28.3 \\
       PandaGPT~\cite{su2023pandagpt} & Vincuna-13B & ImageBind ViT-H/14 & 42.5 & 23.1 & 61.5 & 34.1 & 32.7 & 28.7 & 57.6 \\
       VisualGLM~\cite{du2022glm} & ChatGLM-6B & EVA-CLIP & 33.5 & 11.4 & 48.8 & 27.7 & 35.8 & 17.6 & 41.5 \\
       InstructBLIP~\cite{dai2023instructblip} & Vicuna-7B & EVA-G & 33.9 & 21.6 & 47.4 & 22.5 & 33.0 & 24.4 & 41.1 \\
       LLaVA-v1.5~\cite{liu2023llava1_5} & Vicuna-v1.5-7B & CLIP ViT-L/14 & 59.5 & 32.4 & 72.6 & 49.3 & 62.3 & 52.2 & 67.7 \\
       Otter-I~\cite{li2023otter} & LLaMA-7B & CLIP ViT-L/14 & 48.3 & 22.2 & 63.3 & 39.4 & 46.8 & 36.4 & 60.6 \\
       Shikra~\cite{chen2023shikra} & Vincuna-7B & CLIP ViT-L/14 & 60.2 & 33.5 & 69.6 & 53.1 & 61.8 & 50.4 & 71.7 \\
       LMEye~\cite{li2023lmeye} & FlanT5-XL & CLIP ViT-L/14 & 62.6 & 41.0 & 74.3 & 55.9 & 61.6 & 58.7 & 69.2 \\
       MMICL~\cite{zhao2023mmicl} & FlanT5-XXL & EVA-G & 65.2 & 44.3 & 77.9 & 64.8 & 66.5 & 53.6 & 70.6 \\
       mPLUG-Owl2~\cite{xu2023mplug2} & LLaMA2-7B & CLIP ViT-L/14 & 66.0 & 43.4 & 76.0 & 62.1 & 68.6 & 55.9 & 73.0 \\
       LLaVA-v1.5-13B & Vicuna-v1.5-13B & CLIP ViT-L/14 & 67.8 & 43.4 & 71.9 & 60.7 & 73.4 & 59.1 & 77.3 \\
       \hline
       \rowcolor[HTML]{faf0e6}
       \textbf{LION} & FlanT5-XXL & EVA-G & \textbf{73.4} & \textbf{51.7} & \textbf{84.1} & \textbf{78.4} & \textbf{74.0} & \textbf{60.8} & \textbf{78.9} \\
       \bottomrule[1pt]
    \end{tabular}
    \label{tab:mmbench}
\end{table*}

\begin{figure*}[h!]
    \centering
    \includegraphics[width=0.98\linewidth]{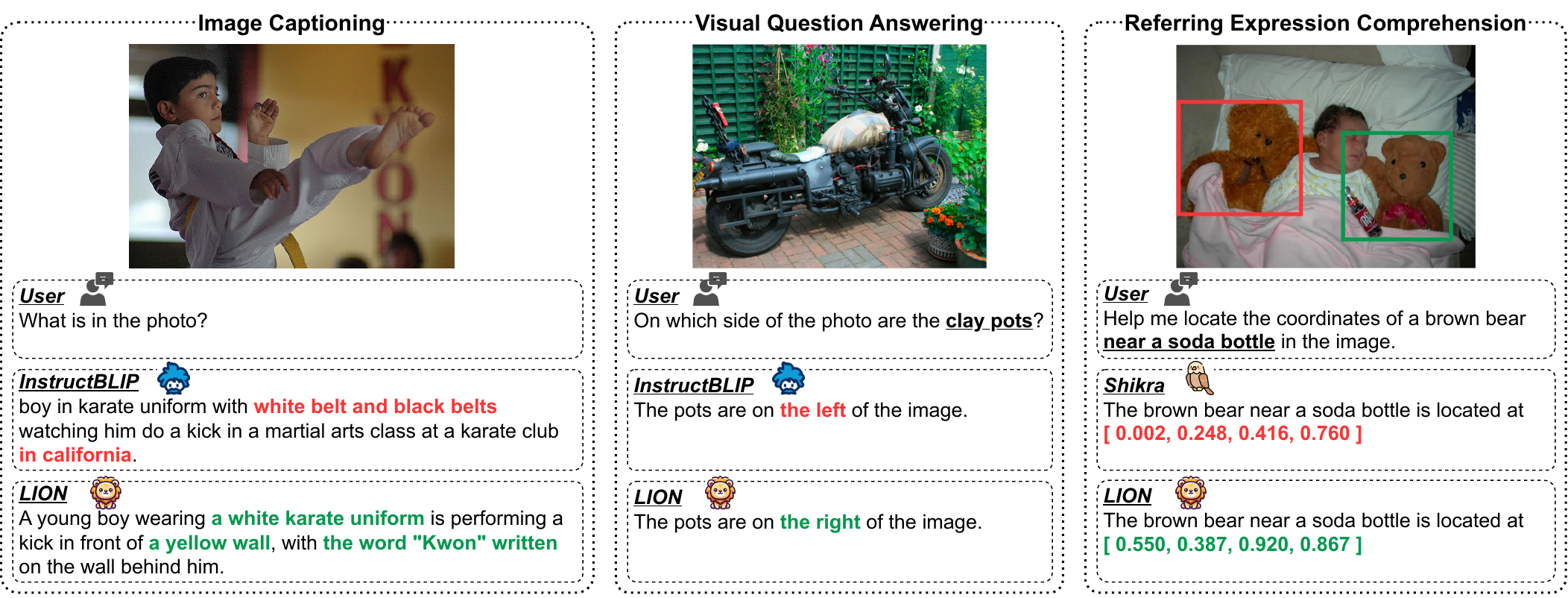}
    \caption{Qualitative comparison of InstructBLIP \includegraphics[width=8pt]{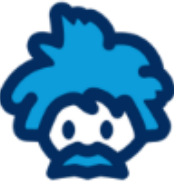}, Shikra \includegraphics[width=8pt]{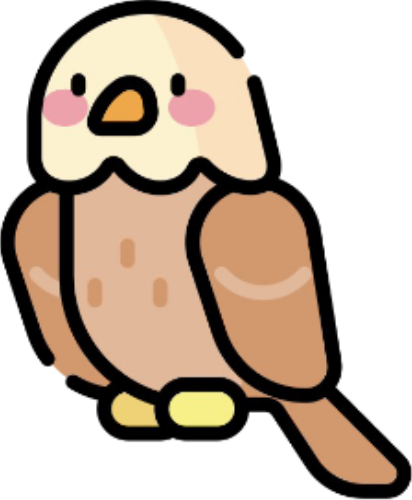}, and LION \includegraphics[width=8pt]{LION_logo.png}. We mark the hallucination or incorrect part in \textcolor[HTML]{FF3333}{red}, and highlight the correct part in \textcolor[HTML]{00994D}{green} for comparison. These samples exhibit that LION is able to achieve superior fine-grained understanding and visual spatial reasoning capabilities with fewer hallucinated responses. }
	\label{fig:qualitative_ana}
\end{figure*}

\subsection{Ablation Study}

\textbf{The effect of vision aggregator.} We conduct an ablation study of the vision aggregator with only visual grounding tasks on LION-4B during stage 2 of the stage-wise instruction-tuning. As illustrated in Fig.~\ref{fig:effect_of_aggregator}, the removal of the vision aggregator degrades REC performances, validating that aggregating multi-level vision features promotes the extraction of fine-grained spatial-aware visual knowledge.

\noindent \textbf{Stage-wise instruction-tuning mitigates the conflict between image-level and region-level tasks.} 
We investigate the performance of two types of VL tasks under three instruction-tuning strategies, i.e., single stage, stage-wise, and stage-wise with a router.
As shown in Table \ref{tab:comp_stage_training}, the stage-wise instruction-tuning strategy shows a significant improvement on the average REC performance, which is completely damaged in the single stage instruction-tuning. The worse REC performance of the single stage strategy can be attributed to the lack of pretraining on large-scale grounding datasets, like in Kosmos-2 and Shikra, and the gap of their input-output modes. To address these challenges, stage-wise training progressively integrates fine-grained spatial-aware knowledge from visual grounding datasets by splitting the whole instruction-tuning process into three stages. The model can sufficiently learn diverse levels of visual knowledge in separate training stages (stages 1 and 2), and incorporate them in the final training stage (stage 3 in Fig.~\ref{fig:stagewise_train}). This contributes to the performance improvements of all VL tasks. Furthermore, stage-wise instruction-tuning with the router improves the held-out and REC performance, with a slight degradation in the held-in performance. All these results demonstrate LION's ability to handle the potential conflict of various VL tasks and maximize the learning benefit.
\noindent \textbf{Dual-level visual knowledge enhances multimodal understanding abilities of MLLMs.} We evaluate the performance of our model integrated with different levels of visual knowledge on various benchmarks in Table \ref{tab:ablation_study}. It can be seen that dual-level visual knowledge can upgrade the performance of all VL tasks to varying degrees. When progressively incorporating fine-grained spatial-aware knowledge, the performances of four tasks, OKVQA, GQA, IconQA, and Visual Spatial Reasoning, are significantly improved, as they highly require region-level understanding and spatial reasoning. When inserting tags as high-level visual evidence, we can see substantial performance increases on Flickr30K and AOKVQA, which demand more comprehensive semantic knowledge than other tasks, like COCO caption and OKVQA.

\noindent \subsection{Evaluations on Object Hallucination and MMBench}
Li \textit{et al.} \cite{li2023evaluating} present an open-sourced evaluation benchmark, called POPE, to evaluate the object hallucination \cite{rawte2023survey}. We follow the POPE evaluation pipeline to inspect LION. The results in Table \ref{tab:obj_hallucination} show that LION has superior results, especially under popular and adversarial settings, which means that incorporating fine-grained and high-level semantic visual knowledge into MLLM can mitigate the object hallucination to some degree.
To comprehensively validate the effectiveness of our method, we further evaluate LION on MMBench \cite{liu2023mmbench}. The results are summarized in Table~\ref{tab:mmbench}. Our strong performances across various skills demonstrate that the progressive incorporation of fine-grained knowledge significantly alleviates the hallucination phenomenon in MLLMs.

\subsection{Qualitative Analysis}
As shown in Fig.~\ref{fig:qualitative_ana}, we depict various examples to validate the advanced perceptual and reasoning abilities of LION. The left example exhibits our superior fine-grained understanding capability to correctly generate the right attributes ``white", ``yellow" and the character ``Kwon". The middle example validates the advantage of our model in visual spatial reasoning. The right example shows that LION accurately localizes the referring object, while Shikra provides an incorrect response caused by the misunderstanding of fine-grained details ``a soda bottle".

\section{Conclusion}

To address the insufficient extraction and reasoning of visual information in MLLMs, we propose LION to exploit dual-level visual knowledge, \textit{i.e.,} fine-grained spatial-aware visual knowledge and high-level semantic visual evidence, based on region-level and image-level VL tasks. To mitigate the internal conflict between these two kinds of tasks, LION proposes a stage-wise instruction-tuning strategy to progressively incorporate fine-grained spatial-aware visual knowledge into MLLMs, achieving the mutual promotion between these two kinds of VL tasks. We use image tags as high-level semantic visual evidence, and present a soft prompting method to alleviate the potential influence resulting from incorrect tags. Extensive experiments validate the superiority of LION in image captioning, VQA, and visual grounding tasks.

{
    \small
    \bibliographystyle{ieeenat_fullname}
    \bibliography{ref}

\begin{thebibliography}{54}
\providecommand{\natexlab}[1]{#1}
\providecommand{\url}[1]{\texttt{#1}}
\expandafter\ifx\csname urlstyle\endcsname\relax
  \providecommand{\doi}[1]{doi: #1}\else
  \providecommand{\doi}{doi: \begingroup \urlstyle{rm}\Url}\fi

\bibitem[Alayrac et~al.(2022)Alayrac, Donahue, Luc, Miech, Barr, Hasson, Lenc,
  Mensch, Millican, Reynolds, et~al.]{alayrac2022flamingo}
Jean-Baptiste Alayrac, Jeff Donahue, Pauline Luc, Antoine Miech, Iain Barr,
  Yana Hasson, Karel Lenc, Arthur Mensch, Katherine Millican, Malcolm Reynolds,
  et~al.
\newblock Flamingo: a visual language model for few-shot learning.
\newblock In \emph{NeurIPS}, 2022.

\bibitem[Chen et~al.(2023{\natexlab{a}})Chen, Qin, Luo, Mi, Li, Sun, and
  Liu]{chen2023pvit}
Chi Chen, Ruoyu Qin, Fuwen Luo, Xiaoyue Mi, Peng Li, Maosong Sun, and Yang Liu.
\newblock Position-enhanced visual instruction tuning for multimodal large
  language models.
\newblock \emph{arXiv preprint arXiv:2308.13437}, 2023{\natexlab{a}}.

\bibitem[Chen et~al.(2023{\natexlab{b}})Chen, Zhu, Shen, Li, Liu, Zhang,
  Krishnamoorthi, Chandra, Xiong, and Elhoseiny]{chen2023minigptV2}
Jun Chen, Deyao Zhu, Xiaoqian Shen, Xiang Li, Zechun Liu, Pengchuan Zhang,
  Raghuraman Krishnamoorthi, Vikas Chandra, Yunyang Xiong, and Mohamed
  Elhoseiny.
\newblock Minigpt-v2: large language model as a unified interface for
  vision-language multi-task learning.
\newblock \emph{arXiv preprint arXiv:2310.09478}, 2023{\natexlab{b}}.

\bibitem[Chen et~al.(2023{\natexlab{c}})Chen, Zhang, Zeng, Zhang, Zhu, and
  Zhao]{chen2023shikra}
Keqin Chen, Zhao Zhang, Weili Zeng, Richong Zhang, Feng Zhu, and Rui Zhao.
\newblock Shikra: Unleashing multimodal llm's referential dialogue magic.
\newblock \emph{arXiv preprint arXiv:2306.15195}, 2023{\natexlab{c}}.

\bibitem[Chen et~al.(2022)Chen, Ge, Tong, Wang, Song, Wang, and
  Luo]{chen2022adaptformer}
Shoufa Chen, Chongjian Ge, Zhan Tong, Jiangliu Wang, Yibing Song, Jue Wang, and
  Ping Luo.
\newblock Adaptformer: Adapting vision transformers for scalable visual
  recognition.
\newblock \emph{Advances in Neural Information Processing Systems},
  35:\penalty0 16664--16678, 2022.

\bibitem[Chen et~al.(2015)Chen, Fang, Lin, Vedantam, Gupta, Doll{\'a}r, and
  Zitnick]{chen2015cococap}
Xinlei Chen, Hao Fang, Tsung-Yi Lin, Ramakrishna Vedantam, Saurabh Gupta, Piotr
  Doll{\'a}r, and C~Lawrence Zitnick.
\newblock Microsoft coco captions: Data collection and evaluation server.
\newblock \emph{arXiv preprint arXiv:1504.00325}, 2015.

\bibitem[Dai et~al.(2023)Dai, Li, Li, Tiong, Zhao, Wang, Li, Fung, and
  Hoi]{dai2023instructblip}
Wenliang Dai, Junnan Li, Dongxu Li, Anthony Meng~Huat Tiong, Junqi Zhao,
  Weisheng Wang, Boyang Li, Pascale Fung, and Steven Hoi.
\newblock Instructblip: Towards general-purpose vision-language models with
  instruction tuning.
\newblock In \emph{NeurIPS}, 2023.

\bibitem[Das et~al.(2017)Das, Kottur, Gupta, Singh, Yadav, Moura, Parikh, and
  Batra]{das2017visdial}
Abhishek Das, Satwik Kottur, Khushi Gupta, Avi Singh, Deshraj Yadav,
  Jos{\'e}~MF Moura, Devi Parikh, and Dhruv Batra.
\newblock Visual dialog.
\newblock In \emph{Proceedings of the IEEE conference on computer vision and
  pattern recognition}, pages 326--335, 2017.

\bibitem[Devlin et~al.(2018)Devlin, Chang, Lee, and Toutanova]{devlin2018bert}
Jacob Devlin, Ming-Wei Chang, Kenton Lee, and Kristina Toutanova.
\newblock Bert: Pre-training of deep bidirectional transformers for language
  understanding.
\newblock \emph{arXiv preprint arXiv:1810.04805}, 2018.

\bibitem[Dosovitskiy et~al.(2020)Dosovitskiy, Beyer, Kolesnikov, Weissenborn,
  Zhai, Unterthiner, Dehghani, Minderer, Heigold, Gelly,
  et~al.]{dosovitskiy2020ViT}
Alexey Dosovitskiy, Lucas Beyer, Alexander Kolesnikov, Dirk Weissenborn,
  Xiaohua Zhai, Thomas Unterthiner, Mostafa Dehghani, Matthias Minderer, Georg
  Heigold, Sylvain Gelly, et~al.
\newblock An image is worth 16x16 words: Transformers for image recognition at
  scale.
\newblock \emph{arXiv preprint arXiv:2010.11929}, 2020.

\bibitem[Du et~al.(2022)Du, Qian, Liu, Ding, Qiu, Yang, and Tang]{du2022glm}
Zhengxiao Du, Yujie Qian, Xiao Liu, Ming Ding, Jiezhong Qiu, Zhilin Yang, and
  Jie Tang.
\newblock Glm: General language model pretraining with autoregressive blank
  infilling.
\newblock In \emph{Proceedings of the 60th Annual Meeting of the Association
  for Computational Linguistics (Volume 1: Long Papers)}, pages 320--335, 2022.

\bibitem[Fang et~al.(2023)Fang, Wang, Xie, Sun, Wu, Wang, Huang, Wang, and
  Cao]{fang2023eva}
Yuxin Fang, Wen Wang, Binhui Xie, Quan Sun, Ledell Wu, Xinggang Wang, Tiejun
  Huang, Xinlong Wang, and Yue Cao.
\newblock Eva: Exploring the limits of masked visual representation learning at
  scale.
\newblock In \emph{CVPR}, 2023.

\bibitem[Gao et~al.(2023)Gao, Han, Zhang, Lin, Geng, Zhou, Zhang, Lu, He, Yue,
  et~al.]{gao2023llamaAdapterV2}
Peng Gao, Jiaming Han, Renrui Zhang, Ziyi Lin, Shijie Geng, Aojun Zhou, Wei
  Zhang, Pan Lu, Conghui He, Xiangyu Yue, et~al.
\newblock Llama-adapter v2: Parameter-efficient visual instruction model.
\newblock \emph{arXiv preprint arXiv:2304.15010}, 2023.

\bibitem[Hendrycks and Gimpel(2016)]{hendrycks2016gaussian}
Dan Hendrycks and Kevin Gimpel.
\newblock Gaussian error linear units (gelus).
\newblock \emph{arXiv preprint arXiv:1606.08415}, 2016.

\bibitem[Hu et~al.(2023)Hu, Xu, Li, Li, Chen, and Tu]{hu2023bliva}
Wenbo Hu, Yifan Xu, Y Li, W Li, Z Chen, and Z Tu.
\newblock Bliva: A simple multimodal llm for better handling of text-rich
  visual questions.
\newblock \emph{arXiv preprint arXiv:2308.09936}, 2023.

\bibitem[Huang et~al.(2023)Huang, Dong, Wang, Hao, Singhal, Ma, Lv, Cui,
  Mohammed, Liu, et~al.]{huang2023kosmos1}
Shaohan Huang, Li Dong, Wenhui Wang, Yaru Hao, Saksham Singhal, Shuming Ma,
  Tengchao Lv, Lei Cui, Owais~Khan Mohammed, Qiang Liu, et~al.
\newblock Language is not all you need: Aligning perception with language
  models.
\newblock \emph{arXiv preprint arXiv:2302.14045}, 2023.

\bibitem[Hudson and Manning(2019)]{hudson2019gqa}
Drew~A Hudson and Christopher~D Manning.
\newblock Gqa: A new dataset for real-world visual reasoning and compositional
  question answering.
\newblock In \emph{Proceedings of the IEEE/CVF conference on computer vision
  and pattern recognition}, pages 6700--6709, 2019.

\bibitem[Jin et~al.(2023)Jin, Mukherjee, Cheng, Shen, Chen, Awadallah, Jose,
  and Ren]{jin2023grill}
Woojeong Jin, Subhabrata Mukherjee, Yu Cheng, Yelong Shen, Weizhu Chen,
  Ahmed~Hassan Awadallah, Damien Jose, and Xiang Ren.
\newblock Grill: Grounded vision-language pre-training via aligning text and
  image regions.
\newblock \emph{arXiv preprint arXiv:2305.14676}, 2023.

\bibitem[Kazemzadeh et~al.(2014)Kazemzadeh, Ordonez, Matten, and
  Berg]{kazemzadeh2014referitgame}
Sahar Kazemzadeh, Vicente Ordonez, Mark Matten, and Tamara Berg.
\newblock Referitgame: Referring to objects in photographs of natural scenes.
\newblock In \emph{Proceedings of the 2014 conference on empirical methods in
  natural language processing (EMNLP)}, pages 787--798, 2014.

\bibitem[Krishna et~al.(2017{\natexlab{a}})Krishna, Zhu, Groth, Johnson, Hata,
  Kravitz, Chen, Kalantidis, Li, Shamma, et~al.]{krishna2017visual}
Ranjay Krishna, Yuke Zhu, Oliver Groth, Justin Johnson, Kenji Hata, Joshua
  Kravitz, Stephanie Chen, Yannis Kalantidis, Li-Jia Li, David~A Shamma, et~al.
\newblock Visual genome: Connecting language and vision using crowdsourced
  dense image annotations.
\newblock \emph{International journal of computer vision}, 123:\penalty0
  32--73, 2017{\natexlab{a}}.

\bibitem[Krishna et~al.(2017{\natexlab{b}})Krishna, Zhu, Groth, Johnson, Hata,
  Kravitz, Chen, Kalantidis, Li, Shamma, et~al.]{krishna2017visualgenome}
Ranjay Krishna, Yuke Zhu, Oliver Groth, Justin Johnson, Kenji Hata, Joshua
  Kravitz, Stephanie Chen, Yannis Kalantidis, Li-Jia Li, David~A Shamma, et~al.
\newblock Visual genome: Connecting language and vision using crowdsourced
  dense image annotations.
\newblock \emph{International journal of computer vision}, 123:\penalty0
  32--73, 2017{\natexlab{b}}.

\bibitem[Li et~al.(2023{\natexlab{a}})Li, Zhang, Chen, Wang, Yang, and
  Liu]{li2023otter}
Bo Li, Yuanhan Zhang, Liangyu Chen, Jinghao Wang, Jingkang Yang, and Ziwei Liu.
\newblock Otter: A multi-modal model with in-context instruction tuning.
\newblock \emph{arXiv preprint arXiv:2305.03726}, 2023{\natexlab{a}}.

\bibitem[Li et~al.(2023{\natexlab{b}})Li, Li, Savarese, and Hoi]{li2023blip2}
Junnan Li, Dongxu Li, Silvio Savarese, and Steven Hoi.
\newblock Blip-2: Bootstrapping language-image pre-training with frozen image
  encoders and large language models.
\newblock In \emph{ICML}, 2023{\natexlab{b}}.

\bibitem[Li et~al.(2022)Li, Zhang, Zhang, Yang, Li, Zhong, Wang, Yuan, Zhang,
  Hwang, et~al.]{li2022grounded}
Liunian~Harold Li, Pengchuan Zhang, Haotian Zhang, Jianwei Yang, Chunyuan Li,
  Yiwu Zhong, Lijuan Wang, Lu Yuan, Lei Zhang, Jenq-Neng Hwang, et~al.
\newblock Grounded language-image pre-training.
\newblock In \emph{CVPR}, 2022.

\bibitem[Li et~al.(2023{\natexlab{c}})Li, Du, Zhou, Wang, Zhao, and
  Wen]{li2023evaluating}
Yifan Li, Yifan Du, Kun Zhou, Jinpeng Wang, Wayne~Xin Zhao, and Ji-Rong Wen.
\newblock Evaluating object hallucination in large vision-language models.
\newblock \emph{arXiv preprint arXiv:2305.10355}, 2023{\natexlab{c}}.

\bibitem[Li et~al.(2023{\natexlab{d}})Li, Hu, Chen, Ma, and Zhang]{li2023lmeye}
Yunxin Li, Baotian Hu, Xinyu Chen, Lin Ma, and Min Zhang.
\newblock Lmeye: An interactive perception network for large language models.
\newblock \emph{arXiv preprint arXiv:2305.03701}, 2023{\natexlab{d}}.

\bibitem[Liu et~al.(2023{\natexlab{a}})Liu, Emerson, and Collier]{liu2023vsr}
Fangyu Liu, Guy Emerson, and Nigel Collier.
\newblock Visual spatial reasoning.
\newblock \emph{Transactions of the Association for Computational Linguistics},
  11:\penalty0 635--651, 2023{\natexlab{a}}.

\bibitem[Liu et~al.(2023{\natexlab{b}})Liu, Li, Li, and Lee]{liu2023llava1_5}
Haotian Liu, Chunyuan Li, Yuheng Li, and Yong~Jae Lee.
\newblock Improved baselines with visual instruction tuning.
\newblock \emph{arXiv preprint arXiv:2310.03744}, 2023{\natexlab{b}}.

\bibitem[Liu et~al.(2023{\natexlab{c}})Liu, Li, Wu, and Lee]{liu2023llava}
Haotian Liu, Chunyuan Li, Qingyang Wu, and Yong~Jae Lee.
\newblock Visual instruction tuning.
\newblock In \emph{NeurIPS}, 2023{\natexlab{c}}.

\bibitem[Liu et~al.(2023{\natexlab{d}})Liu, Duan, Zhang, Li, Zhang, Zhao, Yuan,
  Wang, He, Liu, et~al.]{liu2023mmbench}
Yuan Liu, Haodong Duan, Yuanhan Zhang, Bo Li, Songyang Zhang, Wangbo Zhao, Yike
  Yuan, Jiaqi Wang, Conghui He, Ziwei Liu, et~al.
\newblock Mmbench: Is your multi-modal model an all-around player?
\newblock \emph{arXiv preprint arXiv:2307.06281}, 2023{\natexlab{d}}.

\bibitem[Loshchilov and Hutter(2019)]{loshchilov2017decoupled}
Ilya Loshchilov and Frank Hutter.
\newblock Decoupled weight decay regularization.
\newblock In \emph{ICLR}, 2019.

\bibitem[Lu et~al.(2021)Lu, Qiu, Chen, Xia, Zhao, Zhang, Yu, Liang, and
  Zhu]{lu2021iconqa}
Pan Lu, Liang Qiu, Jiaqi Chen, Tony Xia, Yizhou Zhao, Wei Zhang, Zhou Yu,
  Xiaodan Liang, and Song-Chun Zhu.
\newblock Iconqa: A new benchmark for abstract diagram understanding and visual
  language reasoning.
\newblock \emph{arXiv preprint arXiv:2110.13214}, 2021.

\bibitem[Mao et~al.(2016)Mao, Huang, Toshev, Camburu, Yuille, and
  Murphy]{mao2016refcocog}
Junhua Mao, Jonathan Huang, Alexander Toshev, Oana Camburu, Alan~L Yuille, and
  Kevin Murphy.
\newblock Generation and comprehension of unambiguous object descriptions.
\newblock In \emph{Proceedings of the IEEE conference on computer vision and
  pattern recognition}, pages 11--20, 2016.

\bibitem[Marino et~al.(2019)Marino, Rastegari, Farhadi, and
  Mottaghi]{marino2019okvqa}
Kenneth Marino, Mohammad Rastegari, Ali Farhadi, and Roozbeh Mottaghi.
\newblock Ok-vqa: A visual question answering benchmark requiring external
  knowledge.
\newblock In \emph{Proceedings of the IEEE/cvf conference on computer vision
  and pattern recognition}, pages 3195--3204, 2019.

\bibitem[Peng et~al.(2023)Peng, Wang, Dong, Hao, Huang, Ma, and
  Wei]{peng2023kosmos2}
Zhiliang Peng, Wenhui Wang, Li Dong, Yaru Hao, Shaohan Huang, Shuming Ma, and
  Furu Wei.
\newblock Kosmos-2: Grounding multimodal large language models to the world.
\newblock \emph{arXiv preprint arXiv:2306.14824}, 2023.

\bibitem[Radford et~al.(2021)Radford, Kim, Hallacy, Ramesh, Goh, Agarwal,
  Sastry, Askell, Mishkin, Clark, et~al.]{radford2021CLIP}
Alec Radford, Jong~Wook Kim, Chris Hallacy, Aditya Ramesh, Gabriel Goh,
  Sandhini Agarwal, Girish Sastry, Amanda Askell, Pamela Mishkin, Jack Clark,
  et~al.
\newblock Learning transferable visual models from natural language
  supervision.
\newblock In \emph{International conference on machine learning}, pages
  8748--8763. PMLR, 2021.

\bibitem[Rawte et~al.(2023)Rawte, Sheth, and Das]{rawte2023survey}
Vipula Rawte, Amit Sheth, and Amitava Das.
\newblock A survey of hallucination in large foundation models.
\newblock \emph{arXiv preprint arXiv:2309.05922}, 2023.

\bibitem[Schwenk et~al.(2022)Schwenk, Khandelwal, Clark, Marino, and
  Mottaghi]{schwenk2022aokvqa}
Dustin Schwenk, Apoorv Khandelwal, Christopher Clark, Kenneth Marino, and
  Roozbeh Mottaghi.
\newblock A-okvqa: A benchmark for visual question answering using world
  knowledge.
\newblock In \emph{European Conference on Computer Vision}, pages 146--162.
  Springer, 2022.

\bibitem[Sidorov et~al.(2020)Sidorov, Hu, Rohrbach, and
  Singh]{sidorov2020textcaps}
Oleksii Sidorov, Ronghang Hu, Marcus Rohrbach, and Amanpreet Singh.
\newblock Textcaps: a dataset for image captioning with reading comprehension.
\newblock In \emph{Computer Vision--ECCV 2020: 16th European Conference,
  Glasgow, UK, August 23--28, 2020, Proceedings, Part II 16}, pages 742--758.
  Springer, 2020.

\bibitem[Su et~al.(2023)Su, Lan, Li, Xu, Wang, and Cai]{su2023pandagpt}
Yixuan Su, Tian Lan, Huayang Li, Jialu Xu, Yan Wang, and Deng Cai.
\newblock Pandagpt: One model to instruction-follow them all.
\newblock \emph{arXiv preprint arXiv:2305.16355}, 2023.

\bibitem[Wang et~al.(2023{\natexlab{a}})Wang, Ge, Ding, Kankanhalli, and
  Shan]{wang2023makes}
Guangzhi Wang, Yixiao Ge, Xiaohan Ding, Mohan Kankanhalli, and Ying Shan.
\newblock What makes for good visual tokenizers for large language models?
\newblock \emph{arXiv preprint arXiv:2305.12223}, 2023{\natexlab{a}}.

\bibitem[Wang et~al.(2022)Wang, Yang, Men, Lin, Bai, Li, Ma, Zhou, Zhou, and
  Yang]{wang2022ofa}
Peng Wang, An Yang, Rui Men, Junyang Lin, Shuai Bai, Zhikang Li, Jianxin Ma,
  Chang Zhou, Jingren Zhou, and Hongxia Yang.
\newblock Ofa: Unifying architectures, tasks, and modalities through a simple
  sequence-to-sequence learning framework.
\newblock In \emph{International Conference on Machine Learning}, pages
  23318--23340. PMLR, 2022.

\bibitem[Wang et~al.(2023{\natexlab{b}})Wang, Chen, Chen, Wu, Zhu, Zeng, Luo,
  Lu, Zhou, Qiao, et~al.]{wang2023visionllm}
Wenhai Wang, Zhe Chen, Xiaokang Chen, Jiannan Wu, Xizhou Zhu, Gang Zeng, Ping
  Luo, Tong Lu, Jie Zhou, Yu Qiao, et~al.
\newblock Visionllm: Large language model is also an open-ended decoder for
  vision-centric tasks.
\newblock \emph{arXiv preprint arXiv:2305.11175}, 2023{\natexlab{b}}.

\bibitem[Xu et~al.(2023)Xu, Ye, Yan, Shi, Ye, Xu, Li, Bi, Qian, Wang,
  et~al.]{xu2023mplug2}
Haiyang Xu, Qinghao Ye, Ming Yan, Yaya Shi, Jiabo Ye, Yuanhong Xu, Chenliang
  Li, Bin Bi, Qi Qian, Wei Wang, et~al.
\newblock mplug-2: A modularized multi-modal foundation model across text,
  image and video.
\newblock \emph{arXiv preprint arXiv:2302.00402}, 2023.

\bibitem[Xuan et~al.(2023)Xuan, Guo, Yang, and Zhang]{xuan2023pink}
Shiyu Xuan, Qingpei Guo, Ming Yang, and Shiliang Zhang.
\newblock Pink: Unveiling the power of referential comprehension for
  multi-modal llms.
\newblock \emph{arXiv preprint arXiv:2310.00582}, 2023.

\bibitem[Ye et~al.(2023)Ye, Xu, Xu, Ye, Yan, Zhou, Wang, Hu, Shi, Shi,
  et~al.]{ye2023mplug}
Qinghao Ye, Haiyang Xu, Guohai Xu, Jiabo Ye, Ming Yan, Yiyang Zhou, Junyang
  Wang, Anwen Hu, Pengcheng Shi, Yaya Shi, et~al.
\newblock mplug-owl: Modularization empowers large language models with
  multimodality.
\newblock \emph{arXiv preprint arXiv:2304.14178}, 2023.

\bibitem[You et~al.(2023)You, Zhang, Gan, Du, Zhang, Wang, Cao, Chang, and
  Yang]{you2023ferret}
Haoxuan You, Haotian Zhang, Zhe Gan, Xianzhi Du, Bowen Zhang, Zirui Wang,
  Liangliang Cao, Shih-Fu Chang, and Yinfei Yang.
\newblock Ferret: Refer and ground anything anywhere at any granularity.
\newblock \emph{arXiv preprint arXiv:2310.07704}, 2023.

\bibitem[Young et~al.(2014)Young, Lai, Hodosh, and
  Hockenmaier]{young2014flickr30k}
Peter Young, Alice Lai, Micah Hodosh, and Julia Hockenmaier.
\newblock From image descriptions to visual denotations: New similarity metrics
  for semantic inference over event descriptions.
\newblock \emph{Transactions of the Association for Computational Linguistics},
  2:\penalty0 67--78, 2014.

\bibitem[Yu et~al.(2016)Yu, Poirson, Yang, Berg, and Berg]{yu2016contextinREC}
Licheng Yu, Patrick Poirson, Shan Yang, Alexander~C Berg, and Tamara~L Berg.
\newblock Modeling context in referring expressions.
\newblock In \emph{ECCV}, pages 69--85. Springer, 2016.

\bibitem[Zhai et~al.(2023)Zhai, Tong, Li, Cai, Qu, Lee, and
  Ma]{zhai2023catastrophicforget}
Yuexiang Zhai, Shengbang Tong, Xiao Li, Mu Cai, Qing Qu, Yong~Jae Lee, and Yi
  Ma.
\newblock Investigating the catastrophic forgetting in multimodal large
  language models.
\newblock \emph{arXiv preprint arXiv:2309.10313}, 2023.

\bibitem[Zhang et~al.(2023{\natexlab{a}})Zhang, Han, Zhou, Hu, Yan, Lu, Li,
  Gao, and Qiao]{zhang2023llamaAdapter}
Renrui Zhang, Jiaming Han, Aojun Zhou, Xiangfei Hu, Shilin Yan, Pan Lu,
  Hongsheng Li, Peng Gao, and Yu Qiao.
\newblock Llama-adapter: Efficient fine-tuning of language models with
  zero-init attention.
\newblock \emph{arXiv preprint arXiv:2303.16199}, 2023{\natexlab{a}}.

\bibitem[Zhang et~al.(2023{\natexlab{b}})Zhang, Huang, Ma, Li, Luo, Xie, Qin,
  Luo, Li, Liu, et~al.]{zhang2023ram}
Youcai Zhang, Xinyu Huang, Jinyu Ma, Zhaoyang Li, Zhaochuan Luo, Yanchun Xie,
  Yuzhuo Qin, Tong Luo, Yaqian Li, Shilong Liu, et~al.
\newblock Recognize anything: A strong image tagging model.
\newblock \emph{arXiv preprint arXiv:2306.03514}, 2023{\natexlab{b}}.

\bibitem[Zhao et~al.(2023)Zhao, Cai, Si, Ma, An, Chen, Liu, Wang, Han, and
  Chang]{zhao2023mmicl}
Haozhe Zhao, Zefan Cai, Shuzheng Si, Xiaojian Ma, Kaikai An, Liang Chen, Zixuan
  Liu, Sheng Wang, Wenjuan Han, and Baobao Chang.
\newblock Mmicl: Empowering vision-language model with multi-modal in-context
  learning.
\newblock \emph{arXiv preprint arXiv:2309.07915}, 2023.

\bibitem[Zhu et~al.(2023)Zhu, Chen, Shen, Li, and Elhoseiny]{zhu2023minigpt4}
Deyao Zhu, Jun Chen, Xiaoqian Shen, Xiang Li, and Mohamed Elhoseiny.
\newblock Minigpt-4: Enhancing vision-language understanding with advanced
  large language models.
\newblock \emph{arXiv preprint arXiv:2304.10592}, 2023.

\end{thebibliography}
}


\appendix

\begin{table*}[b]
    \centering
    \caption{The training datasets used for instruction-tuning.}
    \begin{tabular}{l|l|ccc|c}
    \toprule[1pt]
        Task       & Dataset                        & Stage 1    & Stage 2    & Stage 3    & Data Number \\
        \hline
        Dialogue         & LLaVA-Instruct-150K            & \checkmark &            & \checkmark & 361K \\
        VQA              & OKVQA, A-OKVQA, VQAv2, OCR-VQA & \checkmark &            & \checkmark & 1.3M \\
        VQG              & OKVQA, A-OKVQA, VQAv2          & \checkmark &            & \checkmark & 470K   \\
        Image Captioning          & COCO, TextCaps                 & \checkmark &            & \checkmark & 524K  \\
        REC              & Visual Genome                  &            & \checkmark & \checkmark & 3.6M\\
        REG              & Visual Genome                  &            & \checkmark & \checkmark & 3.6M\\
    \bottomrule[1pt]
    \end{tabular}
    \label{tab:train_data_stat}
\end{table*}

\section{Experimental Details}
\paragraph{Architecture.}

We use the off-the-shelf ViT-G/14 from EVA-CLIP~\cite{fang2023eva} without the last layer as our frozen vision backbone. The vision aggregator consists of two Bert Layers \cite{devlin2018bert} with cross attention in each layer. The output from the Vision Aggregator undergoes a transformation via a two-layer MLP with GeLU~\cite{hendrycks2016gaussian} activation, and is projected into the latent feature space of the LLM. This output is then concatenated with the output from Q-Former and the textual inputs, forming the comprehensive inputs for the LLM.
In the LLM, the hidden dimension of each adapter is set to 64. We implement LION on LLMs with two different size, including FlanT5-XL(3B) and FlanT5-XXL(11B), resulting in LION-4B and LION-12B, respectively.

When incorporating the image tags as high-level semantic visual evidence, we use the recognize anything model (RAM-14M) \cite{zhang2023ram} based on the backbone Swin-Large. All the image tags are generated by using a $384\times 384$ image size and a $0.8$ threshold across 4585 categories in the ram tag list. All other hyperparameters are set the same as in the RAM codebase \footnote{https://github.com/xinyu1205/recognize-anything}.

\paragraph{Training Details.}

Our training process comprises three stages. In Stage 1, we use a batch size of 64 for 10 epochs over 30k steps, with a learning rate starting at 1e-5 and reducing to a minimum of 0. Stage 2 increases the batch size to 256 for another 10 epochs across 60k steps, beginning with a learning rate of 5e-4, which is reduced to a floor of 1e-6; notably, the learning rate for the Vision Aggregator is set to a constant 1e-5. Stage 3 reverts to a batch size of 64 for 10 epochs and 60k steps, with an initial learning rate of 1e-5, descending to a minimum of 0. Throughout all stages, the AdamW~\cite{loshchilov2017decoupled} optimizer is employed with \(\beta_1 = 0.9\), \(\beta_2 = 0.999\), and a weight decay of 0.05. The learning rate is warmed up linearly from 1e-8 across 1000 steps at the beginning of each stage.

\paragraph{Training Data.}

We describe all training datasets in Table \ref{tab:train_data_stat}. In stage 1, a part of LION is trained on image-level VL tasks, including COCO Caption, TextCaps, OKVQA, AOKVQA, VQAv2, OCR-VQA. Sepcifically, we follow InstructBLIP to define a visual question generation (VQG) task, which requires the model to generate a question given an answer. This VQG task is formed by using OKVQA, AOKVQA, and VQAv2 training datasets. We also use a dialogue dataset, LLaVA-Instruct-150K in this stage. In stage2, we use Visual Genome training dataset to construct referring expression comprehension (REC) and referring expression generation (REG) tasks. In final stage, all the mentioned datasets are used to train a unified model, resulting in the LION. We insert image tags in stage 3, firstly generate image tags for all training images, then use them with the soft prompting method.
We provide evaluation metrics in Table \ref{tab:eval_metric}. 

\begin{table*}
    \centering
    \caption{Summary of the evaluation datasets.}
    \begin{tabular}{c|l|l|l}
    \toprule[1pt]
        Task & Dataset & Split & Metric \\ \hline
        \multirow{3}{*}{\shortstack{Image\\Captioning}}
         & Flickr30K & karpathy-test & CIDEr($\uparrow$) \\
         & COCO & karpathy-test & CIDEr($\uparrow$) \\
         & TextCaps & val & CIDEr($\uparrow$) \\ \hline
        \multirow{6}{*}{\shortstack{Visual Question\\Answering}}
         & OKVQA & val & Accuracy($\uparrow$) \\
         & AOKVQA & val & Accuracy($\uparrow$) \\
         & Visual Spatial Reasoning & val & Accuracy($\uparrow$) \\
         & Visual Dialog & val & MRR($\uparrow$) \\
         & IconQA & test & Accuracy($\uparrow$) \\
         & GQA & test-dev & Accuracy($\uparrow$) \\ \hline
        \multirow{3}{*}{\shortstack{Referring Expression\\Comprehension}}
         & RefCOCO & val \& testA \& testB & Accuracy($\uparrow$) \\
         & RefCOCO+ & val \& testA \& testB & Accuracy($\uparrow$) \\
         & RefCOCOg & val \& test & Accuracy($\uparrow$) \\
    \bottomrule[1pt]
    \end{tabular}
    \label{tab:eval_metric}
\end{table*}

\section{Instruction Templates}
\subsection{Task Templates for Instruction-Tuning}
We provide instruction templates for transform image-level and region-level VL tasks into a instruction-tuning format. For image-level VL tasks, we follow the setting in InstructBLIP. For region-level tasks, we use the templates in Shikra, which are generated by GPT-4 with carefully designed instructions. For each task listed in Table \ref{tab:train_instructions}, we only show a few templates.
\begin{table*}
    \centering
    \caption{Examples of instruction templates for various tasks. ``\{expr\}" represents the expression in the REC task. ``\{BBox\}" refers to the bounding box of a user-specified location.}
    \begin{tabular}{c|l}
    \toprule[1pt]
       VQA  & \begin{tabular}[l]{@{}l@{}}\textless{}Image\textgreater{}Given the image, answer the following question with no more than three words. \{Question\}\\ \textless{}Image\textgreater{}Based on the image, respond to this question with a short answer: \{Question\}. Answer:\\ \textless{}Image\textgreater{}Use the provided image to answer the question: \{Question\} Provide your answer as short as possible:\\ \textless{}Image\textgreater{}What is the answer to the following question? "\{Question\}"\\ \textless{}Image\textgreater{}The question "\{Question\}" can be answered using the image. A short answer is\end{tabular} \\
       \hline
       VQG  & \begin{tabular}[l]{@{}l@{}}\textless{}Image\textgreater{}Based on the image, provide a question with the answer: \{Answer\}. Question:\\ \textless{}Image\textgreater{}Given the visual representation, create a question for which the answer is "\{Answer\}".\\ \textless{}Image\textgreater{}From the image provided, craft a question that leads to the reply: \{Answer\}. Question:\\ \textless{}Image\textgreater{}Considering the picture, come up with a question where the answer is: \{Answer\}.\\ \textless{}Image\textgreater{}Taking the image into account, generate an question that has the answer: \{Answer\}. Question:\end{tabular}  \\
       \hline
       \shortstack{Image\\Captioning} & \begin{tabular}[l]{@{}l@{}} \textless{}Image\textgreater{}Can you briefly explain what you see in the image?\\ \textless{}Image\textgreater{}Could you use a few words to describe what you perceive in the photo?\\ \textless{}Image\textgreater{}Please provide a short depiction of the picture.					\\ \textless{}Image\textgreater{}Using language, provide a short account of the image.\\ \textless{}Image\textgreater{}Use a few words to illustrate what is happening in the picture.\end{tabular} \\
       \hline
       REC  & \begin{tabular}[l]{@{}l@{}}\textless{}image\textgreater{}Identify the position of \{expr\} in image and share its coordinates.  \\ \textless{}image\textgreater{}I'd like to request the coordinates of \{expr\} within the photo.  \\ \textless{}image\textgreater{}How can I locate \{expr\} in the image? Please provide the coordinates.  \\ \textless{}image\textgreater{}I am interested in knowing the coordinates of \{expr\} in the picture.\\ \textless{}image\textgreater{}Assist me in locating the position of \{expr\} in the photograph and its bounding box coordinates. \\ \textless{}image\textgreater{}In the image, I need to find \{expr\} and know its coordinates. Can you please help?\end{tabular} \\
       \hline
       REG  & \begin{tabular}[c]{@{}l@{}}\textless{}image\textgreater{}What are the unique characteristics of the rectangular section \{BBox\} in image?\\ \textless{}image\textgreater{}Describe the novel qualities of the selected bounding box \{BBox\} in image.\\ \textless{}image\textgreater{}What sets the chosen region \{BBox\} in image apart from its surroundings?\\ \textless{}image\textgreater{}Provide a one-of-a-kind depiction for the area enclosed by \{BBox\} in image.\\ \textless{}image\textgreater{}How would you portray the unique features of the designated box \{BBox\} in image?\\ \textless{}image\textgreater{}Explain the distinguishing characteristics of the marked bounding box \{BBox\} in image.\end{tabular} \\
    \bottomrule[1pt]
    \end{tabular}
    \label{tab:train_instructions}
\end{table*}

\subsection{Instructions for Evaluation}
We provide instructions for evaluation on various benchmarks. For instructions involving options, we arrange the options in the alphabetical order. For REC tasks, we randomly choose a template in training instruction lists for evaluation, which is the same as Shikra.

\noindent\textbf{OKVQA, AOKVQA, GQA} $<$Image$>$ Question: \{Question\} Short answer:

\noindent\textbf{COCOCap, Flickr30K, TextCaps} $<$Image$>$ A short image description:

\noindent\textbf{IconQA} $<$Image$>$ \{Question\}

\noindent\textbf{VSR} $<$Image$>$ Based on the image, is this statement true or false? ``\{Question\}" Answer:

\noindent\textbf{Visual Dialog} $<$Image$>$ Dialog history: \{History\}\textbackslash n Question: \{Question\} Short answer:

\end{document}